\documentclass[11pt]{article}
\usepackage{konvens2019}
\usepackage{mathptmx}
\usepackage{amsmath}
\usepackage[scaled=.90]{helvet}
\usepackage{courier}
\usepackage{booktabs}
\usepackage{graphicx}
\usepackage{url}
\usepackage{latexsym}
\usepackage{paralist}
\usepackage{subcaption}
\usepackage{todonotes}

\graphicspath{{./graphics/}}

\definecolor{darkred}{rgb}{.8,0,0}
\definecolor{darkgreen}{rgb}{0,.5,0}

\title{Does {BERT} Make Any Sense? \\{I}nterpretable Word Sense Disambiguation with Contextualized Embeddings}

\author{Gregor Wiedemann\textsuperscript{1} \quad Steffen Remus\textsuperscript{1} \\
\textsuperscript{1}Language Technology Group\\ 
Department of Informatics\\ 
Universit\"{a}t Hamburg, Germany\\
{\tt \{gwiedemann,remus,biemann\}} \\
{\tt @informatik.uni-hamburg.de} \\ \And
Avi Chawla\textsuperscript{2} \quad Chris Biemann\textsuperscript{1} \\
\textsuperscript{2}Indian Institute of Technology (BHU) \\
Varanasi, India \\
{\tt avi.chawla.cse16@iitbhu.ac.in}}

\date{}

\begin{document}
\maketitle
\begin{abstract}
Contextualized word embeddings (CWE) such as provided by ELMo \cite{peters.2018}, Flair NLP \cite{akbik.2018}, or BERT \cite{devlin.2019} are a major recent innovation in NLP. 
CWEs provide semantic vector representations of words depending on their respective context.
Their advantage over static word embeddings has been shown for a number of tasks, such as text classification, sequence tagging, or machine translation. 
Since vectors of the same word type can vary depending on the respective context, they implicitly provide a model for word sense disambiguation (WSD). 
We introduce a simple but effective approach to WSD using a nearest neighbor classification on CWEs.
We compare the performance of different CWE models for the task and can report improvements above the current state of the art for two standard WSD benchmark datasets.
We further show that the pre-trained BERT model is able to place polysemic words into distinct `sense' regions of the embedding space, while ELMo and Flair NLP do not seem to possess this ability.
\end{abstract}

\section{Synonymy and Polysemy of Word Representations}
Lexical semantics is characterized by a high degree of polysemy, i.e.\ the meaning of a word changes depending on the context in which it is currently used \cite{harris:54:dh}.
Word Sense Disambiguation (WSD) is the task to identify the correct sense of the usage of a word from a (usually) fixed inventory of sense identifiers. 
For the English language, WordNet \cite{fellbaum.1998} is the most commonly used sense inventory providing more than $200$K word-sense pairs. 

To train and evaluate WSD systems, a number of shared task datasets have been published in the SemEval workshop series. In the \textit{lexical sample} task \cite{kilgarriff-2001-english,mihalcea-etal-2004-senseval}, a training set and a test set is provided. The relatively large data contains one sense-annotated word per training/test instance. The \textit{all-words} task \cite{edmonds.2001,snyder.2004} only provides a small number of documents as test data where each ambiguous word is annotated with its sense.
To facilitate the comparison of WSD systems, some efforts have been made to provide a comprehensive evaluation framework \cite{raganato.2017}, and to unify all publicly available datasets for the English language \cite{viali.2018}.

WSD systems can be distinguished into three types --- knowledge-based, supervised, and semi-supervised approaches. 
\textit{Knowledge-based} systems utilize language resources such as dictionaries, thesauri and knowledge graphs to infer senses. 
\textit{Supervised} approaches train a machine classifier to predict a sense given the target word and its context based on an annotated training data set. 
\textit{Semi-supervised} approaches extend manually created training sets by large corpora of unlabeled data to improve WSD performance. 
All approaches rely on some way of context representation to predict the correct sense.
Context is typically modeled via dictionary resources linked with senses, or as some feature vector obtained from a machine learning model.

A fundamental assumption in structuralist linguistics is the distinction between signifier and signified as introduced by Ferdinand de Saussure \cite{Saussure.2001} in the early $20^{\text{th}}$ century. 
Computational linguistic approaches, when using character strings as the only representatives for word meaning, implicitly assume identity between signifier and signified. 
Different word senses are simply collapsed into the same string representation. 
In this respect, word counting and dictionary-based approaches to analyze natural language texts have been criticized as pre-Saussurean \cite{Pecheux.1995}.
In contrast, the distributional hypothesis not only states that meaning is dependent on context. It also states that words occurring in the same contexts tend to have a similar meaning \cite{harris:54:dh}.
Hence, a more elegant way of representing meaning has been introduced by using the contexts of a word as an intermediate semantic representation that mediates between signifier and signified.
For this, explicit vector representations, such as TF-IDF, or latent vector representations, with reduced dimensionality, have been widely used.
Latent vector representations of words are commonly called word embeddings.
They are fixed length vector representations, which are supposed to encode semantic properties. 
The seminal neural word embedding model Word2Vec \cite{mikolov.2013}, for instance, can be trained efficiently on billions of sentence contexts to obtain semantic vectors, one for each word type in the vocabulary.
It allows synonymous terms to have similar vector representations that can be used for modeling virtually any downstream NLP task. 
Still, a polysemic term is represented by one single vector only, which represents all of its different senses in a collapsed fashion.

To capture polysemy as well, the idea of word embeddings has been extended to encode word sense embeddings. \newcite{neelakantan:2015:emnlp} first introduced a neural model to learn multiple embeddings for one word depending on different senses. The number of senses can be defined by a given parameter, or derived automatically in a non-paramentric version of the model. However, employing sense embeddings in any downstream NLP task requires a reliable WSD system in an earlier stage to decide how to choose the appropriate embedding from the sense inventory.

Recent efforts to capture polysemy for word embeddings give up on the idea of a fixed word sense inventory.
Contextualized word embeddings (CWE) do not only create one vector representation for each type in the vocabulary, they also they produce distinct vectors for each token in a given context.
The contextualized vector representation is supposed to represent word meaning and context information. 
This enables downstream tasks to actually distinguish the two levels of the signifier and the signified allowing for more realistic modeling of natural language.
The advantage of such contextually embedded token representations compared to static word embeddings has been shown for a number of tasks such as text classification \cite{zampieri.2019} and sequence tagging \cite{akbik.2018}.

\paragraph{Contribution:}%
We show that CWEs can be utilized directly to approach the WSD task due to their nature of providing distinct vector representations for the same token depending on its context.
To learn the semantic capabilities of CWEs, we employ a simple, yet interpretable approach to WSD using a $k$-nearest neighbor classification (kNN) approach. 
We compare the performance of three different CWE models on four standard benchmark datasets.
Our evaluation yields that not all contextualization approaches are equally effective in dealing with polysemy, and that the simple kNN approach suffers severely from sparsity in training datasets.
Yet, by using kNN, we include provenance into our model, which allows to investigate the training sentences that lead to the classifier's decision. 
Thus, we are able to study to what extent polysemy is captured by a specific contextualization approach.
For two datasets, we are able to report new state-of-the-art (SOTA) results.

\section{Related Work}%
%

\subsection{Neural Word Sense Disambiguation}%
Several efforts have been made to induce different vectors for the multiplicity of senses a word can express.
\newcite{bartunovetal:2016}, \newcite{neelakantan:2015:emnlp},  or \newcite{rothe:2015:ACL-IJCNLP} induce so-called sense embeddings in a pre-training fashion.
While \newcite{bartunovetal:2016} induce sense embeddings in an unsupervised way and only fix the maximum number of senses per word, \newcite{rothe:2015:ACL-IJCNLP} require a pre-labeled sense inventory such as WordNet. Then, the sense embeddings are mapped to their corresponding synsets.
Other approaches include the re-use of pre-trained word embeddings in order to induce new sense embeddings \cite{pelevina:2016:RepL4NLP,remus:2018:lrec}.
\newcite{panchenko.2017} then also use induced sense embeddings for the downstream task of WSD.
\newcite{camacho.2018} provide an extensive overview of different word sense modeling approaches.

For WSD, \mbox{(semi-)}supervised approaches with recurrent neural network architectures represent the current state of the art.
Two major approaches were followed. 
First, \newcite{melamud.2016} and \newcite{yuan.2016}, for instance, compute sentence context vectors for ambiguous target words. 
In the prediction phase, they select nearest neighbors of context vectors to determine the target word sense.
\newcite{yuan.2016} also use unlabeled sentences in a semi-supervised label propagation approach to overcome the sparse training data problem of the WSD task.
Second, \newcite{kageback.2016} employ a recurrent neural network to classify sense labels for an ambiguous target word given its surrounding sentence context. In contrast to earlier approaches, which relied on feature engineering  \cite{taghipour-ng-2015-semi}, their architecture only uses pretrained GloVe word embeddings \cite{pennington2014glove} to achieve SOTA results on two English lexical sample datasets.
For the all-words WSD task, \newcite{vial.2018} also employ a recurrent neural network. But instead of single target words, they sequentially classify sense labels for all tokens in a sentence.
They also introduce an approach to collapse the sense vocabulary from WordNet to unambiguous hypersenses, which increases the label to sample ratio for each label, i.e.\ sense identifier.
By training their network on the large sense annotated datasets SemCor~\cite{miller.1993} and the Princeton Annotated Gloss Corpus based on WordNet synset definitions~\cite{fellbaum.1998}, they achieve the highest performance so far on most all-words WSD benchmarks.
A similar architecture with an enhanced sense vocabulary compression was applied in \cite{vial.2019}, but instead of GloVe embeddings, BERT wordpiece embeddings \cite{devlin.2019} are used as input for training. Especially the BERT embeddings further improved the performance yielding new state-of-the-art results.

\subsection{Contextualized Word Embeddings}%
The idea of modeling sentence or context-level semantics together with word-level semantics proved to be a powerful innovation.
For most downstream NLP tasks, CWEs drastically improved the performance of neural architectures compared to static word embeddings. 
However, the contextualization methodologies differ widely. 
We, thus, hypothesize that they are also very different in their ability to capture polysemy.

Like static word embeddings, CWEs are trained on large amounts of unlabeled data by some variant of language modeling.
In our study, we investigate three most prominent and widely applied approaches: \textbf{Flair} \cite{akbik.2018}, \textbf{ELMo} \cite{peters.2018}, and \textbf{BERT} \cite{devlin.2019}. 

\paragraph{Flair:}%
For the contextualization provided in the \textit{Flair NLP} framework, \newcite{akbik.2018} take a static pre-trained word embedding vector, e.g.\ the GloVe word embeddings \cite{pennington2014glove}, and concatenate two context vectors based on the left and right sentence context of the word to it. 
Context vectors are computed by two recurrent neural models, one character language model trained from left to right, one another from right to left. 
Their approach has been applied successfully especially for sequence tagging tasks such as named entity recognition and part-of-speech tagging.

\paragraph{ELMo:}%
\textit{Embeddings from language models} (ELMo) \cite{peters.2018} approaches contextualization similar to Flair, but instead of two character language models, two stacked recurrent models for words are trained, again one left to right, and another right to left. 
For CWEs, outputs from the embedding layer, and the two bidirectional recurrent layers are not concatenated, but collapsed into one layer by a weighted, element-wise summation.

\paragraph{BERT:}%
In contrast to the previous two approaches, \textit{Bidirectional Encoder Representations from Transformers} (BERT) \cite{devlin.2019} does not rely on the merging of two uni-directional recurrent language models with a (static) word embedding, but provides contextualized token embeddings in an end-to-end language model architecture. 
For this, a self-attention based transformer architecture is used, which, in combination with a masked language modeling target, allows to train the model seeing all left and right contexts of a target word at the same time. 
Self-attention and non-directionality of the language modeling task result in extraordinary performance gains compared to previous approaches.

\vspace{1em}
\noindent
According to the distributional hypothesis, if the same word regularly occurs in different, distinct contexts, we may assume polysemy of its meaning \cite{miller1991contextual}.
Contextualized embeddings should be able to capture this property. 
In the following experiments, we investigate this hypothesis on the example of the introduced models.

\begin{table*}[t]
\resizebox{\textwidth}{!}{%
\begin{tabular}{@{}rrrrrrrrr@{}}
  \toprule
  ~ & \textbf{SE-2 (Tr)} & \textbf{SE-2 (Te)} & \textbf{SE-3 (Tr)} & \textbf{SE-3 (Te)} & \textbf{S7-T7 (coarse)} & \textbf{S7-T17 (fine)} & \textbf{SemCor} & \textbf{WNGT} \\
  \midrule
  \#sentences & 8,611 & 4,328 & 7,860 & 3,944 & 126 & 245 & 37,176 & 117,659 \\
  \#CWEs & 8,742 & 4,385 & 9,280 & 4520 & 455 & 6,118 & 230,558 & 1,126,459 \\
  \#distinct words & 313 & 233 & 57 & 57 & 327 & 1,177 & 20,589 & 147,306 \\
  \#senses & 783 & 620 & 285 & 260 & 371 & 3,054 & 33,732 & 206,941\\
  avg \#senses p. word & 2.50 & 2.66 & 5.00 & 4.56 & 1.13 & 2.59 & 1.64 & 1.40 \\
  avg \#CWEs p. word \& sense & 11.16 & 7.07 & 32.56 & 17.38 & 1.23 & 2.00 & 6.83 & 5.44\\
  avg $k'$ & 2.75 & - & 7.63 & - & - & - & 3.16 & 2.98 \\
  \bottomrule
\end{tabular}%
}
\caption{Properties of our datasets. For the test sets (Te), we do not report $k'$ since they are not used as kNN training instances.}\label{tab:datasets}
\end{table*}

\section{Nearest Neighbor Classification for WSD}
We employ a rather simple approach to WSD using non-parametric nearest neighbor classification (kNN) to investigate the semantic capabilities of contextualized word embeddings.
Compared to parametric classification approaches such as support vector machines or neural models, kNN has the advantage that we can directly investigate the training examples that lead to a certain classifier decision.

The kNN classification algorithm \cite{cover.1967} assigns a plurality vote of a sample's nearest labeled neighbors in its vicinity.
In the most simple case, one-nearest neighbor, it predicts the label from the nearest training instance by some defined distance metric.
Although complex weighting schemes for kNN exist, we stick to the simple non-parametric version of the algorithm to be able to better investigate the semantic properties of different contextualized embedding approaches.

As distance measure for kNN, we rely on cosine distance of the CWE vectors.
Our approach considers only senses for a target word that have been observed during training.
We call this approach localized nearest neighbor word sense disambiguation.
We use spaCy\footnote{\small\url{https://spacy.io/}} \cite{spacy:2015:EMNLP} for pre-processing and the lemma of a word as the target word representation, e.g.\ `danced', `dances' and `dancing' are mapped to the same lemma `dance'.
Since BERT uses wordpieces, i.e.\ subword units of words instead of entire words or lemmas, we re-tokenize the lemmatized sentence and average all wordpiece CWEs that belong to the target word.
Moreover, for the experiments with BERT embeddings\footnote{We use the \texttt{bert-large-uncased} model.}, we follow the heuristic by \newcite{devlin.2019} and concatenate the averaged wordpiece vectors of the last four layers.

We test different values for our single hyper-parameter $k \in \lbrace 1, \dots, 10, 50, 100, 500, 1000 \rbrace$. 
Like words in natural language, word senses follow a power-law distribution.
Due to this, simple baseline approaches for WSD like the \textit{most frequent sense (MFS)} baseline are rather high and hard to beat.
Another effect of the skewed distribution are imbalanced training sets.
Many senses described in WordNet only have one or two example sentences in the training sets, or are not present at all.
This is severely problematic for larger $k$ and the default implementation of kNN because of the majority class dominating the classification result.
To deal with sense distribution imbalance, we modify the majority voting of kNN to $k' = min(k, \vert V_s \vert)$ where $V_s$ is the set of CWEs with the least frequent training examples for a given word sense $s$.

\section{Datasets}%
We conduct our experiments with the help of four standard WSD evaluation sets, two lexical sample tasks and two all-words tasks. 
As lexical sample tasks, SensEval-2 \cite[SE-2]{kilgarriff-2001-english} and SensEval-3 \cite[SE-3]{mihalcea-etal-2004-senseval} provide a training data set and test set each.
The all-words tasks of SemEval~2007 Task~7 \cite[S7-T7]{navigli.2007} and Task~17 \cite[S7-T17]{pradhan.2007} solely comprise test data, both with a substantial overlap of their documents. 
The two sets differ in granularity: While ambiguous terms in Task~17 are annotated with one WordNet sense only, in Task~7 annotations are coarser clusters of highly similar WordNet senses.
For training of the all-words tasks, we use %
\begin{inparaenum}[\itshape a\upshape)]
\item the SemCor dataset \cite{miller.1993}, and
\item the Princeton WordNet gloss corpus (WNGT) \cite{fellbaum.1998}
\end{inparaenum}
separately to investigate the influence of different training sets on our approach.
For all experiments, we utilize the suggested datasets as provided by the UFSAC framework\footnote{Unification of Sense Annotated Corpora and Tools. We are using Version 2.1: \small\url{https://github.com/getalp/UFSAC}} \cite{viali.2018}, i.e.\ the respective training data.
A concise overview of the data can be found in Table~\ref{tab:datasets}.
From this, we can observe that the SE-2 and SE-3 training data sets, which were published along with the respective test sets, provide many more examples per word and sense than SemCor or WNGT.

\section{Experimental Results}\label{sec:results}%
We conduct two experiments to determine whether contextualized word embeddings can solve the WSD task. 
In our first experiment, we compare different pre-trained embeddings with $k=1$.
In our second experiment, we test multiple values of $k$ and the BERT pre-trained embeddings\footnote{BERT performed best in experiment one.} in order to estimate an optimal $k$.
Further, we qualitatively examine the results to analyze, which cases can be typically solved by our approach and where it fails.

\subsection{Contextualized Embeddings}%
\begin{table}[t]
\centering
\resizebox{0.48\textwidth}{!}{%
\begin{tabular}{@{}rrrrrrr@{}}
\toprule
\textbf{Model} & \textbf{SE-2} & \textbf{SE-3} & \multicolumn{2}{l}{\textbf{S7-T7 (coarse)}} & \multicolumn{2}{l}{\textbf{S7-T17 (fine)}} \\
& & & {\small SemCor} & {\small WNGT} & {\small SemCor} & {\small WNGT} \\
\midrule
Flair          & 65.27         & 68.75         & 69.24            & 78.68       & 45.92        & 50.99           \\
ELMo           & 67.57         & 70.70         & 70.80            & 79.12       & 52.61        & 50.11           \\
BERT           & \textbf{76.10}         & \textbf{78.62}         & 73.61     & \textbf{81.11}              & \textbf{59.82}      & 55.16             \\ \bottomrule
\end{tabular}%
}
\caption{kNN with $k=1$ WSD performance (F1\%). Best results for each testset are marked bold.}\label{tab:one-nn}
\end{table}
%
To compare different CWE approaches, we use $k=1$ nearest neighbor classification.
Table~\ref{tab:one-nn} shows a high variance in performance. 
Simple kNN with ELMo as well as BERT embeddings beats the state of the art of the lexical sample task SE-2 (cp. Table~3). BERT also outperforms all others on the SE-3 task.

However, we observe a major performance drop of our approach for the two all-words WSD tasks in which no training data is provided along with the test set.
For S7-T7 and S7-T17, the content and structure of the out-of-domain SemCor and WNGT training datasets differ drastically from those in the test data, which prevents yielding near state-of-the-art results.
In fact, similarity of contextualized embeddings largely relies on semantically \emph{and} structurally similar sentence contexts of polysemic target words. 
Hence, the more example sentences can be used for a sense, the higher are the chances that a nearest neighbor expresses the same sense.
As can be seen in Table \ref{tab:datasets}, the SE-2 and SE-3 training datasets provide more CWEs for each word and sense, and our approach performs better with a growing number of CWEs, even with a higher average number of senses per word as is the case in SE-3. 

Thus, we conclude that the nearest neighbor approach suffers specifically from data sparseness.
The chances increase that aspects of similarity other than the sense of the target word in two compared sentence contexts drive the kNN decision.
Moreover, CWEs actually do not organize well in spherically shaped form in the embedding space.
Although senses might be actually separable, the non-parametric kNN classification is unable to learn a complex decision boundary focusing only on the most informative aspects of the CWE \cite[p.~4]{yuan.2016}.

\subsection{Nearest Neighbors}

\begin{table}
\centering
\resizebox{0.48\textwidth}{!}{%
\begin{tabular}{@{}rrrrr@{}}
\toprule
\textbf{k} & \textbf{SE-2} & \textbf{SE-3} & \textbf{S7-T7} & \textbf{S7-T17} \\ \midrule
1          & 76.10                & 78.62                 & 81.11       & 59.82                   \\
2          & 76.10                & 78.62                 & 81.11       & 59.82                   \\
3          & \underline{\textbf{76.52}}    & 79.66                 & 80.94                   & 59.38                   \\
4          & \underline{\textbf{76.52}}    & 79.55                 & 80.94                   & 59.82                   \\
5          & 76.43                & 79.79                 & 81.07                   & 60.27                   \\
6          & 76.43                & 79.81                 & 81.07                   & 60.27                   \\
7          & 76.50                & 80.02                 & 81.03                   & 60.49                   \\
8          & 76.50                & 79.86                 & 81.03                   & 60.49                   \\
9          & 76.40                & 79.97                 & 81.03                   & 60.49                   \\
10         & 76.40                & \underline{\textbf{80.12}}     & 81.03                   & 60.49                   \\ 
50 & 76.43 & 79.66 & 81.11 & \underline{60.94} \\
100 & 76.43 & 79.63 & \underline{81.20} & 60.71 \\
500 & 76.38 & 79.63 & 81.11 & 60.71 \\
1000 & 76.38 & 79.63 & 81.11 & 60.71 \\
\midrule
MFS & 54.79 & 58.95 & 70.94 & 48.44 \\
K{\aa}geb{\"a}ck (2016) & \textit{66.90}          & \textit{73.40}              & -                       & -                       \\
\newcite{yuan.2016}  & -                & -                 & 84.30                   & 63.70                   \\
\newcite{vial.2018}  & -                & -                 & 86.02                   & 66.81                   \\
\newcite{vial.2019}  & -                & -                 & \textit{\textbf{90.60}}                   & \textit{\textbf{71.40}}                   \\ \bottomrule
\end{tabular}%
}
\caption{Best kNN models vs.\ most frequent sense (MFS) and state of the art (F1\%). Best results are bold, previous SOTA is in italics and our best results are underlined.}\label{tab:knn}
\end{table}

\paragraph{\textit{K}-Optimization:}%
To attenuate for noise in the training data, we optimize for $k$ to obtain a more robust nearest neighbor classification. 
Table~\ref{tab:knn} shows our best results using the BERT embeddings along with results from related works.
For \mbox{SensEval-2} and \mbox{SensEval-3}, we achieve a new state-of-the-art result. 
We observe convergence with higher $k$ values since our $k'$ normalization heuristic is activated.
For the S7-T*, we also achieve minor improvements with a higher $k$, but still drastically lack behind the state of the art.

\begin{figure*}[ht]%
\centering
\resizebox{\textwidth}{!}{
\begin{subfigure}[b]{.32\linewidth}
\centering
\fbox{\includegraphics[width=\textwidth, clip, trim={160 160 150 160}]{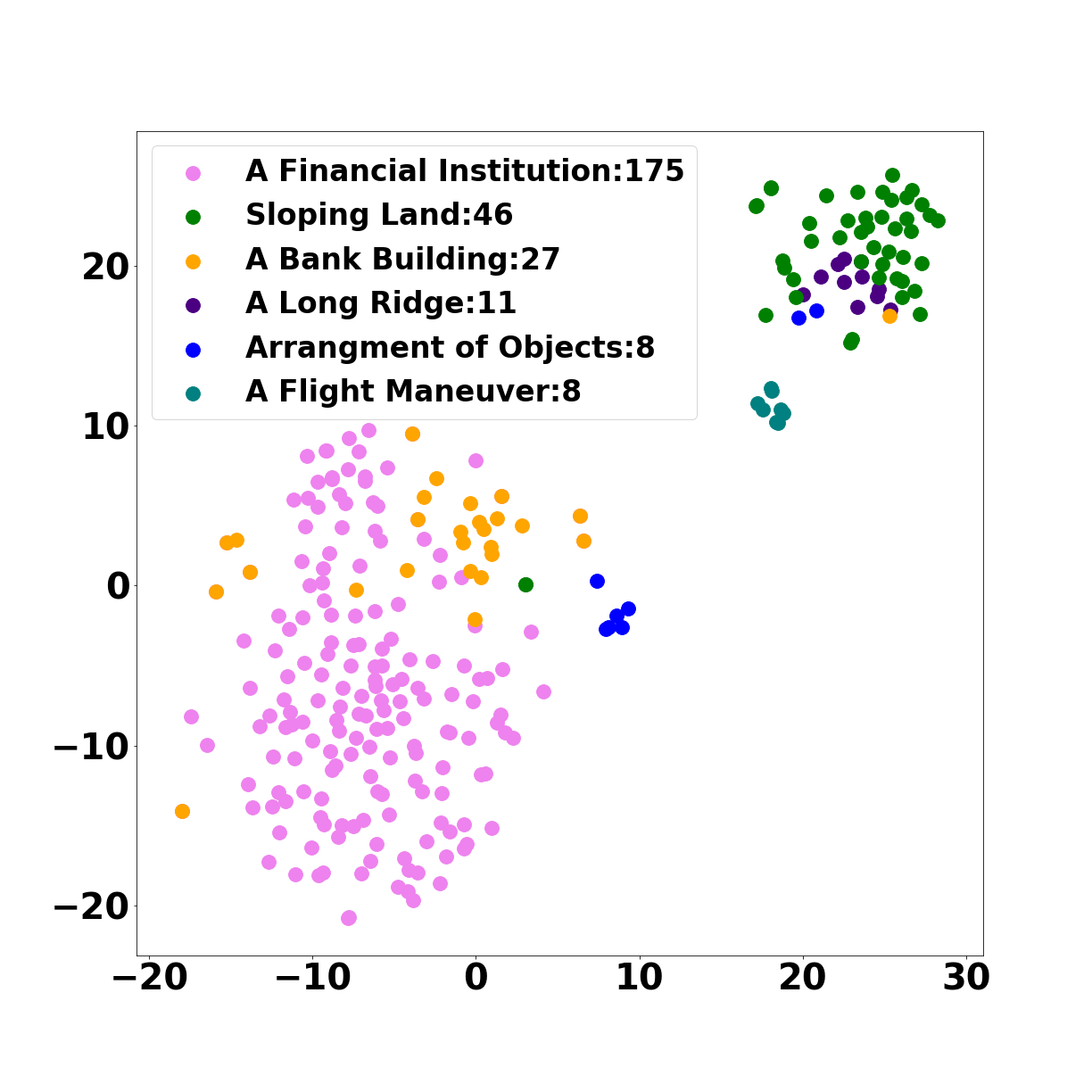}}
\subcaption{BERT}
\end{subfigure}
~
\begin{subfigure}[b]{.32\linewidth}
\centering
\fbox{\includegraphics[width=\textwidth, clip, trim={160 160 150 160}]{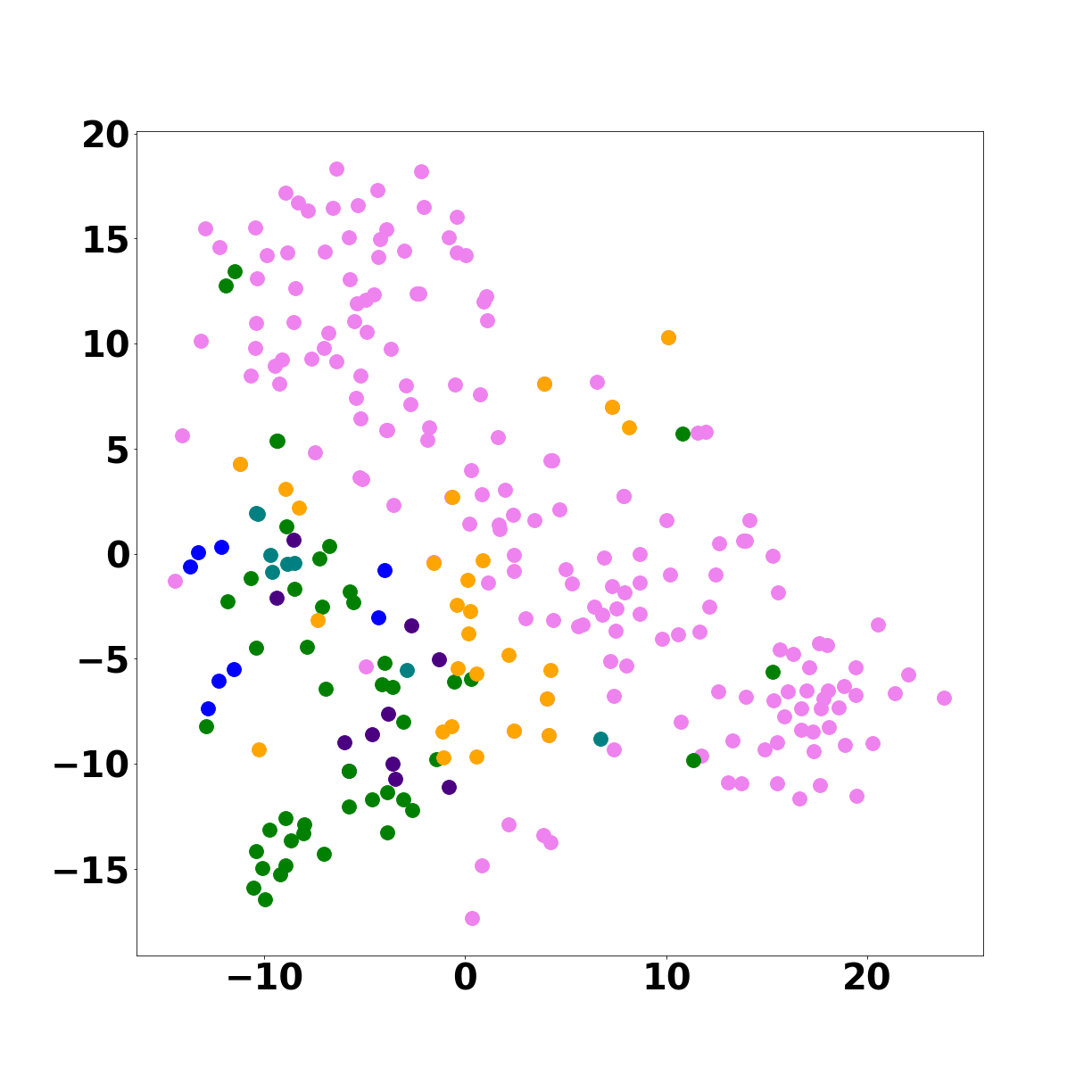}}
\subcaption{Flair}
\end{subfigure}
~
\begin{subfigure}[b]{.32\linewidth}
\centering
\fbox{\includegraphics[width=\textwidth, clip, trim={160 160 150 160}]{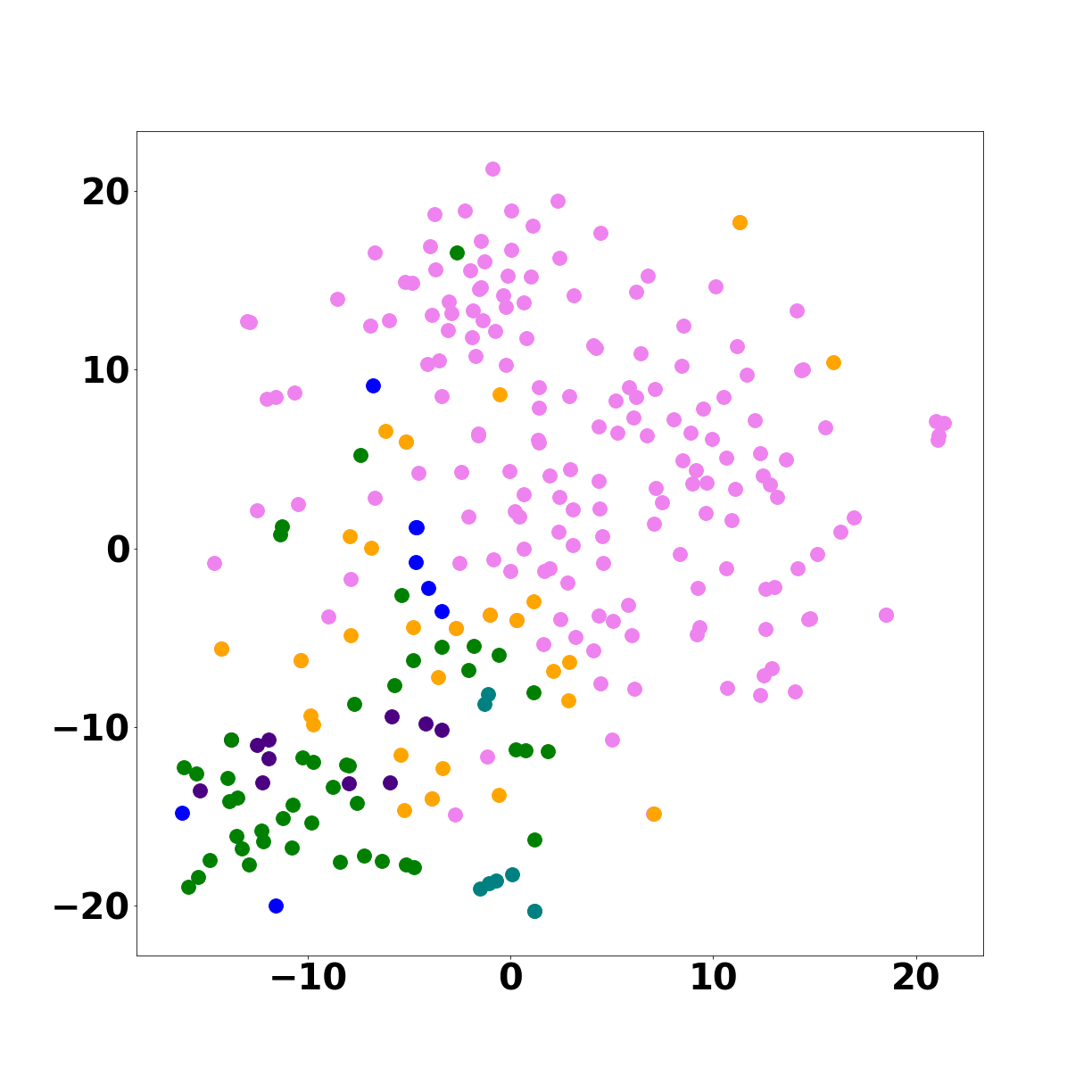}}
\subcaption{ELMo}
\end{subfigure}
}
\caption{T-SNE plots of different senses of `bank' and their contextualized embeddings. The legend shows a short description of the respective WordNet sense and the frequency of occurrence in the training data. Here, the SE-3 training dataset is used.}
\label{fig:tsne}
\end{figure*}
%
\paragraph{Senses in CWE space:}%
We investigate how well different CWE models encode information such as distinguishable senses in their vector space.
Figure~\ref{fig:tsne} shows T-SNE plots~\cite{vanDerMaaten2008} of six different senses of the word ``bank'' in the \mbox{SE-3} training dataset encoded by the three different CWE methods. For visualization purposes, we exclude senses with a frequency of less than two.
The Flair embeddings hardly allow to distinguish any clusters as most senses are scattered across the entire plot. 
In the ELMo embedding space, the major senses are slightly more separated in different regions of the point cloud.
Only in the BERT embedding space, some senses form clearly separable clusters.
Also within the larger clusters, single senses are spread mostly in separated regions of the cluster.
Hence, we conclude that BERT embeddings actually seem to encode some form of sense knowledge, which also explains why kNN can be successfully applied to them. 
Moreover, we can see why a more powerful parametric classification approach such as employed by \newcite{vial.2019} is able to learn clear decision boundaries.
Such clear decision boundaries seem to successfully solve the data sparseness issue of kNN.

\paragraph{Error analysis:}%
From a qualitative inspection of true positive and false positive predictions, we are able to infer some semantic properties of the BERT embedding space and the used training corpora.
Table~\ref{tab:nn_sample} shows selected examples of polysemic words in different test sets, including their nearest neighbor from the respective training set.

\begin{table*}[]
\centering
\resizebox{\textwidth}{!}{%
{\footnotesize
\begin{tabular}{@{}rp{0.43\textwidth}p{0.43\textwidth}@{}}
\toprule
& ~\hfill~\textbf{Example sentence}~\hfill~ & ~\hfill~\textbf{Nearest neighbor}~\hfill~   \\ \midrule
& ~\hfill~SE-3 (train)~\hfill~ & ~\hfill~SE-3 (test)~\hfill~ \\\midrule
(1) & President Aquino, admitting that the death of Ferdinand Marcos had sparked a wave of sympathy for the late dictator, urged Filipinos to stop weeping for the man who had laughed all the way to the \textbf{bank}\textsubscript{[A Bank Building]}. \vspace{0.15cm} & 
They must have been filled in at the \textcolor{darkgreen}{\textbf{bank}\textsubscript{[A Bank Building]}} either by Mr Hatton himself or else by the cashier who was attending to him. \\
(2) & Soon after setting off we came to a forested valley along the \textbf{banks}\textsubscript{[Sloping Land]} of the Gwaun. &
In my own garden the twisted hazel, corylus avellana contorta, is underplanted with primroses, bluebells and wood anemones, for that is how I remember them growing, as they still do, along the \textcolor{darkgreen}{\textbf{banks}\textsubscript{[Sloping Land]}} of the rive Greta \vspace{0.15cm}  \\
(3) & In one direction only a little earthy \textbf{bank}\textsubscript{[A Long Ridge]} separates me from the edge of the ocean, while in the other the valley goes back for miles and miles. \vspace{.15cm} &
The lake has been swept clean of snow by the wind, the sweepings making a huge \textcolor{darkgreen}{\textbf{bank}\textsubscript{[A Long Ridge]}} on our side that we have to negotiate. \\
(4) & However, it can be possible for the documents to be signed after you have sent a payment by cheque provided that you arrange for us to hold the cheque and not pay it into the \textbf{bank}\textsubscript{[A Financial Institution]} until we have received the signed deed of covenant. \vspace{0.15cm}  &
The purpose of these stubs in a paying -- in book is for the holder to have a record of the amount of money he had deposited in his \textcolor{darkred}{\textbf{bank}\textsubscript{[A Bank Building]}}. \\
(5) & He continued: assuming current market conditions do not deteriorate further, the group, with conservative borrowings, a prime land \textbf{bank}\textsubscript{[A Financial Institution]} and a good forward sales position can look forward to another year of growth. \vspace{0.15cm} & 
Crest Nicholson be the exception, not have much of a land \textcolor{darkred}{\textbf{bank}\textsubscript{[Supply or Stock]}} and rely on its skill in land buying. \\
(6) & The marine said, get down behind that grass \textbf{bank}\textsubscript{[A Long Ridge]}, sir, and he immediately lobbed a mills grenade into the river. & 
The guns were all along the river \textcolor{darkred}{\textbf{bank}\textsubscript{[Sloping Land]}} as far as I could see. \vspace{0.5cm}\\ \toprule
& ~\hfill~SemCor~\hfill~ & 
~\hfill~S7-T17~\hfill~   \\ \midrule
(7) & Some 30 \textbf{balloon}\textsubscript{[Large Tough Nonrigid Bag]} shows are held annually in the U.S., including the world's largest convocation of ersatz Phineas Foggs -- the nine-day Albuquerque International Balloon Fiesta that attracts some $800,000$ enthusiasts and more than 500 balloons, some of which are fetchingly shaped to resemble Carmen Miranda, Garfield or a 12-story-high condom. \vspace{0.15cm} & 
Homes and factories and schools and a big wide federal highway, instead of peaceful corn to rest your eyes on while you tried to rest your heart, while you tried not to look at the \textcolor{darkgreen}{\textbf{balloon}\textsubscript{[Large Tough Nonrigid Bag]}} and the bandstand and the uniforms and the flash of the instruments. \\
(8) & The condom \textbf{balloon}\textsubscript{[Large Tough Nonrigid Bag]} was denied official entry status this year. & 
Just like the \textcolor{darkgreen}{\textbf{balloon}\textsubscript{[Large Tough Nonrigid Bag]}} would go up and you could sit all day and wish it would spring a leak or blow to hell up and burn and nothing like that would happen. \vspace{0.15cm} \\
(9) & Starting with congressman Mario Biaggi (now serving a jail \textbf{sentence}\textsubscript{[The Period of Time a Prisoner Is Imprisoned]}), the company began a career of bribing federal, state and local public officials and those close to public officials, right up to and including E. Robert Wallach, close friend and adviser to former attorney general Ed Meese. \vspace{0.15cm} & 
When authorities convicted him of practicing medicine without a license (he got off with a suspended \textcolor{darkgreen}{\textbf{sentence}\textsubscript{[The Period of Time a Prisoner Is Imprisoned]}} of three years because of his advanced age of 77), one of his victims was not around to testify: he was dead of cancer." \\
(10) & Americans it seems have followed Malcolm Forbes's hot-air lead and taken to \textbf{balloon}\textsubscript{[To Ride in a Hot-Air Balloon]} in a heady way. &
Just like the \textcolor{darkred}{\textbf{balloon}\textsubscript{[Large Tough Nonrigid Bag]}} would go up and you could sit all day and wish it would spring a leak or blow to hell up and burn and nothing like that would happen. \vspace{0.15cm} \\
(11) & Any question as to why an author would believe this plaintive, high-minded note of assurance is necessary is answered by reading this \textbf{book}\textsubscript{[A Published Written Work]} about sticky fingers and sweaty scammers. \vspace{0.15cm} &
But the \textcolor{darkred}{\textbf{book}\textsubscript{[A Written Version of a Play]}} is written around a somewhat dizzy cartoonist, and it has to be that way. \\
(12) & In between came lots of coffee drinking while \textbf{watching}\textsubscript{[To Look Attentively]} the balloons inflate and lots of standing around deciding who would fly in what balloon and in what order [\dots]. &
So Captain Jenks returned to his harbor post to \textcolor{darkred}{\textbf{watch}\textsubscript{[To Follow With the Eyes or the Mind; observe]}} the scouting plane put in five more appearances, and to feel the certainty of this dread rising within him. \\ \bottomrule
\end{tabular}}%
}
\caption{Example predictions based on nearest neighbor sentences. The word in question is marked in boldface, subset with a short description of its WordNet synset (true positives \textcolor{darkgreen}{green}, false positives \textcolor{darkred}{red}).
}\label{tab:nn_sample}
\end{table*}

%
Not only vocabulary overlap in the context as in `along the \textit{bank} of the river' and `along the \textit{bank} of the river Greta' (2) allows for correct predictions, but also semantic overlap as in `little earthy bank' and `huge bank [of snow]' (3).
On the other hand, vocabulary overlap, as well as semantic relatedness as in `land bank' (5) can lead to false predictions. 
Another interesting example for the latter is the confusion between `grass bank' and `river bank' (6) where the nearest neighbor sentence in the training set shares some military context with the target sentence.
The correct sense (\emph{bank\%1:17:01::Sloping Land}) and the predicted sense (\emph{bank\%1:17:00::A Long Ridge or Pile [of earth]}) share high semantic similarity, too. In this example, they might even be used interchangeably.
Apparently this context yields higher similarity than any of the other training sentences containing `grass bank' explicitly.

In Example (10), the targeted sense is an action, i.e.\ a verb sense, while the predicted sense is a noun, i.e.\ a different word class.   
In general, this could be easily fixed by restricting the classifier decision to the desired POS. 
However, while example (12) is still a false positive, it nicely shows that the simple kNN approach is able to distinguish senses by word class even though BERT never learned POS classes explicitly.
This effect has been investigated in-depth by \newcite{jawahar-etal-2019-bert}, who found that each BERT layer learns different structural aspects of natural language.
Example (12) also emphasizes the difficulty of distinguishing verb senses itself, i.e.\ the correct sense label in this example is \emph{watch\%2:39:00::look attentively} whereas the predicted label and the nearest neighbor is \emph{watch\%2:41:00::follow with the eyes or the mind; observe}. Verb senses in WordNet are very fine grained and thus harder to distinguish automatically and by humans, too.

\subsection{Post-evaluation experiment}
\begin{table}
\centering
\resizebox{0.48\textwidth}{!}{%
\begin{tabular}{@{}rrrrrrrr@{}}
\toprule
\textbf{k} & \textbf{SE-2} & \bf SE-3 & \multicolumn{2}{c}{\textbf{S7-T7}} & \multicolumn{2}{c}{\bf S7-T17} \\ 
& & & {\small SemCor} & {\small WNGT} & {\small SemCor} & {\small WNGT} \\
\midrule
1 & 76.10 & 78.62 & 79.30 & 85.23 & 61.38 & 61.98 \\
3 & 76.52 & 79.66 & \bf 79.44 & 85.01 & 60.94 & \bf 62.64 \\
7 & 76.50 & 80.02 & 79.35 & 85.05 & 62.50 & 62.20 \\
10 & 76.40 & 80.12 & 79.40 & 85.10 & 62.72 & 62.20 \\
30 & 76.43 & 79.66 & 79.40 & 85.14 & \bf 63.17 & 61.98 \\
70 & 76.43 & 79.61 & 79.35 & 85.23 & 62.95 & 61.98 \\
100 & 76.43 & 79.63 & 79.35 & \bf 85.32 & 62.95 & 61.98 \\
300 & 76.43 & 79.63 & 79.35 & \bf 85.32 & 62.95 & 61.98 \\
\bottomrule
\end{tabular}%
}
\caption{Best POS-sensitive kNN models. Bold numbers are improved results over Table \ref{tab:knn}.}\label{tab:posteval}
\end{table}

\begin{table}[t]
\centering
\resizebox{0.48\textwidth}{!}{%
\begin{tabular}{@{}lrrrr|r@{}}
\toprule
 & Noun & Adj & Verb & Other & \parbox{.2\linewidth}{avg \#POS per word} \\
\midrule
SE-2 (tr) & 41.32 & 16.57 & 42.11 & 0.00 & 1.0 \\
SE-2 (te) & 40.98 & 16.56 & 42.46 & 0.00 & 1.0 \\
SE-3 (tr) & 46.45 & 3.94 & 49.61 & 0.00 & 1.0 \\
SE-3 (te) & 46.17 & 3.98 & 49.86 & 0.00 & 1.0 \\
S7-T7 & 49.00 & 15.75 & 26.14 & 9.11 & 1.03 \\
S7-T17 & 34.95 & 0.00 & 65.05 & 0.00 & 1.01 \\
SemCor & 38.16 & 14.70 & 38.80 & 8.34 & 1.10 \\
WNGT & 57.93 & 21.57 & 15.55 & 4.96 & 1.07 \\
\bottomrule
\end{tabular}%
}
\caption{Percentage of senses with a certain POS tag in the corpora.}\label{tab:pos}
\end{table}

%
In order to address the issue of mixed POS senses, we run a further experiment, which restricts words to their lemma and their POS tag.  
Table \ref{tab:posteval} shows that including the POS restriction increases the F1 scores for S7-T7 and S7-T17.
This can be explained by the number of different POS tags that can be found in the different corpora (c.f.\ Table \ref{tab:pos}).
The results are more stable with respect to their relative performance, i.e.\ SemCor and WNGT reach comparable scores on S7-T17.
Also, the results for SE-2 and SE-3 did not change drastically.
This can be explained by the average number of POS tags a certain word is labeled with.
This variety is much stronger in the S7-T* tasks compared to SE-*.

\section{Conclusion}
In this paper, we tested the semantic properties of contextualized word embeddings (CWEs) to address word sense disambiguation.%
\footnote{The source code of our experiments is publicly available at: \url{https://github.com/uhh-lt/bert-sense}}
To test their capabilities to distinguish different senses of a particular word, by placing their contextualized vector representation into different regions of the shared vector space, we used a k-nearest neighbor approach, which allows us to investigate their properties on an example basis.  
For experimentation, we used pre-trained models from Flair NLP \cite{akbik.2018}, ELMo \cite{peters.2018}, and BERT \cite{devlin.2019}.
Further, we tested our hypothesis on four standard benchmark datasets for word sense disambiguation.
We conclude that WSD can be surprisingly effective using solely CWEs.
We are even able to report improvements over state-of-the-art results for the two lexical sample tasks of SenseEval-2 and SensEval-3.

Further, experiments showed that CWEs in general are able to capture senses, i.e.\ words, when used in a different sense, are placed in different regions.
This effect appeared strongest using the BERT pre-trained model, where example instances even form clusters.
This might give rise to future directions of investigation, e.g.\ unsupervised word sense-induction using clustering techniques.

\noindent
Since the publication of the BERT model, a number of extensions based on transformer architectures and language model pre-training have been released. In future work, we plan to evaluate also XLM \cite{lample:2019}, RoBERTa \cite{Liu.2019} and XLNet \cite{yang.2019} with our approach. 
In our qualitative error analysis, we observed many near-misses, i.e.\ the target sense and the predicted sense are not particularly far away.
We will investigate if more powerful classification algorithms for WSD based on contextualized embeddings are able to solve this issue even in cases of extremely sparse training data.
%

%

\section*{Acknowledgments}
This work was funded by BWFG Hamburg in the Forum 4.0 project, by DFG in the JOIN-T 2 project and by DAAD in form of a WISE stipend.  
\bibliographystyle{konvens2019}
\bibliography{references}

\end{document}